\documentclass[11pt,a4paper]{article}
\usepackage[hyperref]{acl2021}
\usepackage{times}
\usepackage{latexsym}

\usepackage{microtype}

\usepackage{verbatim}
\usepackage{pifont}
\usepackage{multirow}
\usepackage{amsmath}
\usepackage{capt-of}
\usepackage{tabularx}
\usepackage[caption=false]{subfig}
\usepackage{epsfig}
\usepackage{amssymb}
\usepackage{amsfonts}
\usepackage{booktabs}
\usepackage{scalerel}
\usepackage[inline]{enumitem}
\usepackage{listings}
\usepackage{varwidth}
\usepackage[export]{adjustbox}
\usepackage{tikz}
\usetikzlibrary{tikzmark}
\usepackage{todonotes}
\usepackage{cleveref}

\usepackage{stmaryrd}
\usepackage{bbm}

\usepackage{amsmath,amsfonts,bm}

\def\eqref#1{equation~\ref{#1}}

\def\1{\bm{1}}

\DeclareMathAlphabet{\mathsfit}{\encodingdefault}{\sfdefault}{m}{sl}
\SetMathAlphabet{\mathsfit}{bold}{\encodingdefault}{\sfdefault}{bx}{n}

\usepackage{graphicx}
\usepackage{url}

\usepackage{footnote}
\makesavenoteenv{table}

\aclfinalcopy 

\title{Improving BERT with Syntax-aware Local Attention}
\author{Zhongli Li$^1$\thanks{\; Contribution done during internship at Tencent Cloud Xiaowei.},~~Qingyu Zhou$^2$\thanks{\; Corresponding author.},~~Chao Li$^2$,~~Ke Xu$^1$,~~Yunbo Cao$^2$ \\
  $^1$Beihang University \\
  $^2$Tencent Cloud Xiaowei \\
  \texttt{\{lizhongli@,kexu@nlsde.\}buaa.edu.cn} \\
  \texttt{\{qingyuzhou,diegoli,yunbocao\}@tencent.com}}

\begin{document}
\maketitle

\begin{abstract}
Pre-trained Transformer-based neural language models, such as BERT, have achieved remarkable results on varieties of NLP tasks. 
Recent works have shown that attention-based models can benefit from more focused attention over local regions. Most of them restrict the attention scope within a linear span, or confine to certain tasks such as machine translation and question answering.
In this paper, we propose a syntax-aware local attention, where the attention scopes are restrained based on the distances in the syntactic structure.
The proposed syntax-aware local attention can be integrated with pretrained language models, such as BERT, to render the model to focus on syntactically relevant words.
We conduct experiments\footnote{The code is available at \url{https://github.com/Neutralzz/syntax_aware_local_attention}} on various single-sentence benchmarks, including sentence classification and sequence labeling tasks. 
Experimental results show consistent gains over BERT on all benchmark datasets. 
The extensive studies verify that our model achieves better performance owing to more focused attention over syntactically relevant words.

\end{abstract}

\section{Introduction}

Recently, Transformer~\citep{transformer} has performed remarkably well, standing on the multi-headed dot-product attention which fully takes into account the global contextualized information. Several studies find that self-attention can be enhanced by local attention, where the attention scopes are restricted to important local regions.
\citet{luong,yang-a,xu-leveraging,diff-window} utilize dynamic or fixed windows to perform local attention.
\citet{lisa,sgnet,dep-sa-nmt} explore to utilize syntax to restrain attention for better performance, but each of them confines to a certain task.

In this work, we propose a \textbf{s}yntax-aware \textbf{l}ocal \textbf{a}ttention (\textbf{SLA}) which is adaptable to several tasks, and integrate it with BERT~\citep{bert}.
We first apply dependency parsing to the input text, and calculate the distances of input words to construct the self-attention masks. The local attention scores are calculated by applying these masks to the dot-product attention. 
Then we incorporate the syntax-aware local attention with the Transformer global attention. A gate unit is employed for each token in each layer, which determines how much attention is paid to syntactically relevant words.
We lift weights from existing pre-trained BERT, and evaluate our models on several single-sentence benchmarks, including sentence classification and sequence labeling tasks. Experimental results show that our method achieves consistent performance gains over BERT and outperforms previous syntax-based approaches on the average performance. Furthermore, we compare our syntax-aware local attention with the window-based local attention. We find that the syntax-aware local attention is more involved in the aggregation of local and global attention. The attention visualization also validates the syntactic information supports to capture important local regions.

To summarize, this paper makes the following contributions:
 i) SLA can capture the information of important local regions on the syntactic structure.
 ii) SLA can be easily integrated to Transformer, which allows initialization from pre-trained BERT by increasing very few parameters.
 iii) Experiments show the effectiveness of SLA on various single-sentence benchmarks.

\begin{figure*}[t]
     \centering
     \includegraphics[width=14.5cm]{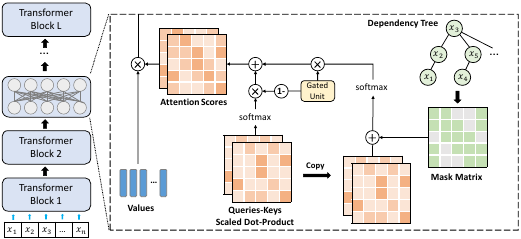}
     \caption{An overview of our model.}
     \label{fig:method}
\end{figure*}

\section{Related Work}
\subsection{Transformer Attention}
Transformer~\citep{transformer} use stacked self-attentions to encode contextual information for input tokens.
The calculation of self-attention depends on the three components of queries $\mathbf{Q}$, keys $\mathbf{K}$ and values $\mathbf{V}$, which are projected from the hidden vectors of the previous layer. 
Then the attention output $\mathbf{A}$ of one head is computed as follows:
\begin{equation}
\begin{split}
    \mathbf{M}_{ij} & = \left\{
        \begin{array}{cl}
            0, & \text{allow to attend} \\
            -\infty, & \text{prevent from attending}
        \end{array}
        \right. \\
    \mathbf{A} & = \text{softmax}(\frac{\mathbf{QK}^T}{\sqrt{d}}+\mathbf{M})\mathbf{V} 
\end{split}
\end{equation}
where $d$ is the dimension of keys and the mask matrix $\mathbf{M}$ controls whether two tokens can attend each other. Within the standard self-attention layer, global attention mechanism is employed that each token provides information to other tokens in the input sentence.

\subsection{Local Attentions}
Local attention involves limiting each token to attend to a subset of the other tokens in the input. Many works utilize a fixed or dynamic window to derive the important local regions.
\citet{luong} first propose a Gaussian-based local attention and increase BLEU scores for neural machine translation. \citet{yang-a} improve the method of \citet{luong} by predicting a central position and window size to model localness. Compared with \citet{yang-a}, \citet{diff-window} attempt to derive the local window span by a soft-masking method. 
However, \citet{levy2014dependency} suggest that more informative representations can be learned from the syntactic structure, instead of a window of surrounding tokens. \citet{lisa} propose to train one attention head to attend to each token's syntactic parent for semantic role labeling. \citet{sgnet} also leverage the syntactic information to self-attention, but confine to question answering. Thus, we explore to take advantage of the syntactic structure to improve the model performance on various benchmarks.

\section{Approach}

In this section, we first introduce the syntax-aware local attention, and then integrate it with standard Transformer attention.
As shown in Figure~\ref{fig:method}, we extend the Transformer layer with the syntax-aware local attention. Syntax-based masking is applied to the dot-product of queries and keys. The final attention scores are computed by incorporating local attention with standard global attention. We stack new layers and initialize weights from pre-trained BERT.

\subsection{Syntax-aware Local Attention}

We derive syntactic structure from dependency parsing, and treat it as an undirected tree. 
Each token $x_i$ is mapped to a tree node $v_i$, and the distance of node $v_i$ and $v_j$ is denoted by $dis(v_i, v_j)$. 
However, the input may be an ungrammatical sentence in some tasks, and the dependency parser is not very accurate. 
Thus, we calculate the distance from neighboring tokens of $x_i$ to token $x_j$ as:
\begin{equation}
    D(i, j) = \text{min}~dis(v_k, v_j),~~~~k \in [i-1, i+1]
\end{equation}
The motivation is that many attention heads specialize in attending heavily on the next or previous token~\citep{bert-look}.
Then, in order to determine whether token $x_j$ can attend to token $x_i$, a threshold $m$ is applied to restrict the distance $D(i, j)$.
For simplification, the mask matrix $\mathbf{M}^{loc}$ calculation can be formulated as:

\begin{equation}
    \mathbf{M}^{loc}_{ij} = \left\{
        \begin{array}{cl}
            0, & D(i, j) \le m \\
            -\infty, & otherwise
        \end{array}
        \right. \\
    \label{eq:locm}
\end{equation}

Given the query $\mathbf{Q}$ and key $\mathbf{K}$ projected from the hidden vectors $\mathbf{H}$, the syntax-aware local attention scores $\mathbf{S}^{loc}$ are formally defined as:
\begin{equation}
    \mathbf{S}^{loc} = \text{softmax}(\frac{\mathbf{QK}^T}{\sqrt{d}}+\mathbf{M}^{loc})
    \label{eq:loca}
\end{equation}
where $d$ is the dimension of keys. In this local attention, two tokens can attend to each other only if they are close enough in the dependency tree. 

\subsection{Attention Aggregation}
\label{sec:combine}
As shown in Figure~\ref{fig:method}, the final attention is the aggregation of syntax-aware local attention and Transformer attention. We denote the Transformer attention scores by $\mathbf{S}^{glb}$. A gated unit is used to combine the global and local attention scores. The gate value  $g_i$ for each token $x_i$ is calculated as follows:
\begin{equation}
    g_i = \sigma (\mathbf{W}_g h_i + b_g),
\end{equation}
where $h_i$ is the hidden vector of token $x_i$ from the previous layer, $\mathbf{W}_g$ is a learnable linear transformation and $b_g$ is the bias. 
Then the attention output $\hat{\mathbf{A}}_i$ is calculated as a weighted average over values $\mathbf{V}$, and the weights are derived from global and local attention scores:
\begin{equation}
    \hat{\mathbf{A}}_i = (g_i \mathbf{S}^{loc}_i +(1-g_i) \mathbf{S}^{glb}_i) \mathbf{V}.
    \label{eq:comb}
\end{equation}

A larger gate value means more focused attention over syntactically relevant words. It can be seen that, if all the outputs of gated units are equal to $0$, we could obtain the standard Transformer attention.
Compared with the original architecture, our self-attention layer has one more input ($\mathbf{M}_{loc}$) and two more trainable parameters ($\mathbf{W}_g$ and $b_g$). Thus, we can easily lift weights from existing pre-trained BERT models.

\begin{table*}[t]
    \centering
    \small
    \begin{tabular}{llcp{0.66cm}<{\centering}cp{0.66cm}<{\centering}p{0.66cm}<{\centering}p{0.66cm}<{\centering}p{0.66cm}<{\centering}p{0.66cm}<{\centering}p{0.66cm}<{\centering}}
    \toprule
        ~ & ~ & ~ & \textbf{CoLA} & \textbf{SST-2} & \multicolumn{3}{c}{\textbf{CoNLL-2003}} & \multicolumn{3}{c}{\textbf{FCE}~(M2)} \\
        Models & Params & Avg. & MCC & Acc & P & R & F$_{1}$ & P & R & F$_{0.5}$ \\
        \midrule
        \multicolumn{11}{l}{\emph{State-of-the-art Models}} \\
        \quad ERNIE 2.0~\citep{ERNIE2} & - & - & 65.4 & 96.0 & - & - & - & - & - & - \\
        \quad T5~\citep{T5}  & - & - & 71.6 & 97.5 & - & - & - & - & - & - \\
        \quad BERT-MRC~\citep{mrc-ner} & - & - & - & - & 92.3 & 94.6 & 93.0 & - & - & - \\
        \quad BERT-GED~\citep{context_key} & - & - & - & - & - & - & - & 65.0 & 38.9 & 57.3 \\
        \midrule
        \multicolumn{11}{l}{\emph{Base-size Models}} \\
        \quad BERT~\citep{bert} & 110M & - & 58.9 & 92.7 & - & - & \textbf{92.4} & - & - & - \\
        \quad LISA~\citep{lisa} & 110M & 74.8 & 59.8 & 92.0 & 90.7 & 92.2 & 91.4 & \textbf{63.4} & 38.6 & \textbf{56.1} \\
        \quad SGNet~\citep{sgnet} & 133M & 74.8 & 59.2 & 93.1 & 90.9 & 92.6 & 91.7 & 60.9 & 40.7 & 55.4 \\
        \quad BERT (Our reimplementation) & 110M & 74.6 & 58.7 & 93.1 & 91.0 & 92.3 & 91.6 & 60.5 & 40.0 & 54.9 \\
        \quad\quad + WLA & +\,0.01M & 75.0 & 59.6 & 92.8 & 91.3 & 92.9 & 92.1 & 60.4 & \textbf{41.3} & 55.3 \\
        \quad\quad + SLA & +\,0.01M & \textbf{75.3} & ~\textbf{60.0}$^\uparrow$ & \textbf{93.3} & ~\textbf{91.5}$^\uparrow$ & ~\textbf{92.9}$^\uparrow$ & ~92.2$^\uparrow$ & ~61.0$^\uparrow$ & ~41.3$^\uparrow$ & ~55.7$^\uparrow$ \\
        
        \midrule
        \multicolumn{11}{l}{\emph{Large-size Models}} \\
        \quad BERT~\citep{bert} & 340M & - & ~60.6$^*$ & ~~93.2$^*$ & - & - & 92.8 & - & - & - \\
        \quad LISA~\citep{lisa} & 340M & 76.2 & 62.2 & 92.7 & 91.3 & 92.6 & 92.0 & 63.4 & 43.2 & 57.9 \\
        \quad SGNet~\citep{sgnet} & 381M & 76.6 & 63.3 & 93.6 & 91.5 & 92.8 & 92.1 & 63.1 & 42.5 & 57.5 \\
        \quad BERT (Our reimplementation) & 340M & 76.9 & 63.9 & 94.0 & 91.7 & 93.1 & 92.4 & 62.7 & 42.6 & 57.3 \\
        \quad\quad + WLA & +\,0.02M & 76.6 & 62.7 & 93.9 & 91.5 & 93.1 & 92.3 & 61.9 & \textbf{44.5} & 57.4 \\
        \quad\quad + SLA & +\,0.02M & \textbf{77.4} & ~\textbf{64.5}$^\uparrow$ & ~~\textbf{94.3}$^\uparrow$ & ~\textbf{92.3}$^\uparrow$ & \textbf{93.4} & \textbf{92.9} & ~\textbf{63.9}$^\uparrow$ & 42.3 & ~\textbf{58.0}$^\uparrow$ \\
        
    \bottomrule
    \end{tabular}
    \caption{Results on single-sentence benchmarks. Results with ``$*$" are taken from~\citet{roberta}. ``$\uparrow$" means statistically significant improvement over the BERT baseline with p-value $< 0.05$. Reported results are averaged over 5 runs. ``Params" is short for the number of model parameters. ``MCC" is short for the Matthews correlation coefficient.}
    \label{tab:res_en}
\end{table*}

\section{Experiments}


\subsection{Experimental Setup}

\paragraph{Benchmarks} We use two English single-sentence classification datasets from the GLUE benchmark~\cite{wang2018glue}. We test on the CoLA and SST-2 datasets for acceptability and sentiment classification. 
Besides, we evaluate our method on two sequence labeling tasks: named entity recognition (NER) and grammatical error detection (GED). We use the CoNLL-2003 and FCE datasets for NER and GED, respectively. The training procedures are introduced in Appendix~\ref{sec:train}.

\paragraph{Configuration} All the training experiments are based on BERT. We use the uncased version of BERT for CoLA and SST-2, and the cased version for CoNLL-2003 and FCE. We derive dependency tree using Spacy\footnote{\url{https://spacy.io/}}. More implementation details are reported in Appendix~\ref{sec:apd_details}.

\paragraph{Baselines} We apply the syntax-aware local attention (\textbf{SLA}) to BERT. In addition to comparing with BERT, we also investigate the following approaches:

\textbf{SGNet}\quad \citet{sgnet} present a syntax-guided self-attention layer, where each word is limited to interact with all of its syntactic ancestor words. Then they stack this layer on the top of the pre-trained BERT model\footnote{\url{https://github.com/cooelf/SG-Net}}, instead of modifying the Transformer architecture.

\textbf{LISA}\quad \citet{lisa} restrict each token to attend to its syntactic parent in one attention head\footnote{\url{https://github.com/strubell/LISA}}. We apply it to BERT and add the corresponding supervision at the last attention head in each Transformer layer.

Besides, we implement the window-based local attention (\textbf{WLA}), which allows each token to attend to the neighboring tokens within a window size $2k+1$ (varying $k$ in \{3,4,5\}). Then it is also integrated with BERT as shown in Section~\ref{sec:combine}.

\subsection{Main Results}
Experimental results are shown in Table~\ref{tab:res_en}. We report results on the dev set of CoLA and SST-2 and the test set of CoNLL-2003 and FCE. We employ t-tests to see if the mean difference differed from 0 between the standard attention and our proposed attention.
It can be seen that our method achieves consistent gains over BERT on single-sentence classification and sequence labeling tasks. 
Specifically, our model exceeds the published BERT results by 3.9\% correlation coefficient on CoLA and 1.1\% accuracy on SST-2.
For the NER task, even though our reimplementation didn't achieve the performance (92.8 F1) reported by~\citet{bert}, our model still outperforms it in large-size.
More importantly, the syntax-aware local attention yields state-of-the-art results with 0.7 absolute improvements on FCE.

Besides, the proposed local attention outperforms other approaches leveraging syntactic information on the average performance. Compared with BERT, the syntax-aware local attention improves performances consistently but the window-based local attention can't. This suggest that BERT can benefit from more attention over syntactically relevant words on several datasets.

However, there are still some gaps between our model and the state-of-the-art models on these datasets. We argue that our method just modifies the standard Transformer attention without changing its main architecture, but those models are trained by using more advanced pre-training methods~\citep{ERNIE2}, larger-scale datasets~\citep{T5}, or learning framework~\citep{mrc-ner}.

\begin{figure}[t]
     \centering
     \includegraphics[width=7.8cm]{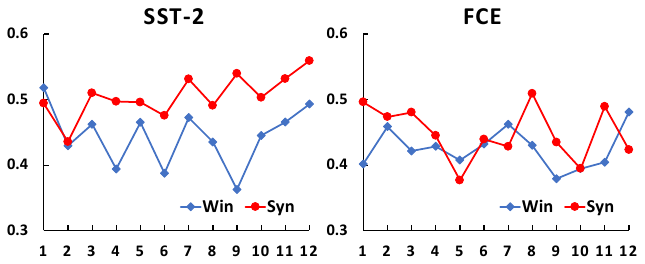}
     \caption{Gate values in different layers on SST-2 and FCE datasets. The blue polyline means that we incorporate the window-based local attention with global attention, and the red polyline corresponds to the syntax-aware local attention.}
     \label{fig:gateunit}
\end{figure}

\begin{figure*}[t]
     \centering
     \includegraphics[width=16.0cm]{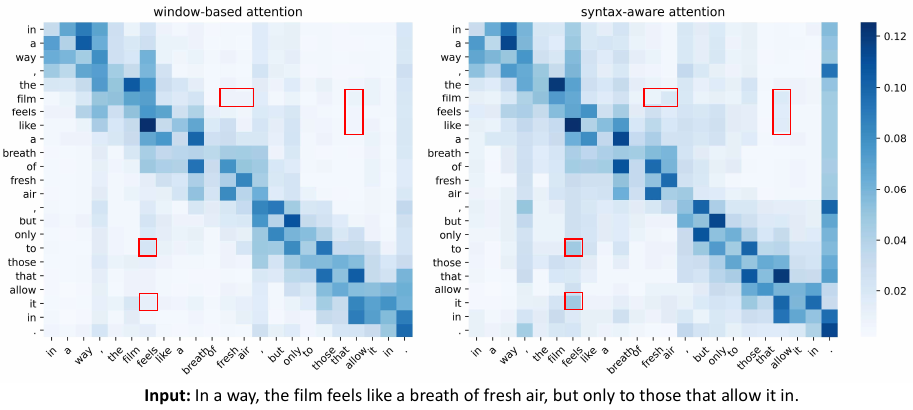}
     \caption{Visualization of attention scores averaged over all heads and all layers. This case is selected from the SST-2 dev set. The red rectangle indicates higher scores on the right side but lower scores on the left side.}
     \label{fig:attention}
\end{figure*}

\subsection{Analysis}

\paragraph{Gated Unit in Each Layer}
It can be seen from Equation~(\ref{eq:comb}) that a larger gate value means a more important role of local attention in the attention aggregation. We analyze the gate values in different layers on SST-2 and FCE datasets. The gated unit outputs are collected from the best-trained base-size models, and are averaged over all input tokens in each layer.

As shown in Figure~\ref{fig:gateunit}, on the SST-2 dataset, the syntax-aware local attention has higher values than the window-based local attention in most layers. Even if the sentences of the FCE dataset are ungrammatical, our attention plays a more important role in 8 of 12 layers.
It indicates that our local attention is more important in the attention score calculation process.
Besides, Table~\ref{tab:res_en} and Figure~\ref{fig:gateunit} illustrate that our model achieves better performances owing to more attention on syntactically relevant words. 

\paragraph{Attention Visualization}
In order to compare the syntax-aware attention with the window-based attention, we plot their attention scores in Figure~\ref{fig:attention}. As formulated in Equation~(\ref{eq:comb}), the attention scores are calculated from the aggregation of global and local attention. We mainly focus on the interactions of tokens, except for \texttt{[CLS]} and \texttt{[SEP]}. Then the attention scores are averaged over all heads and layers.
This visualization validates the effectiveness of incorporating syntactic information into self-attention. As shown in Figure~\ref{fig:attention}, we can see that there are many informative tokens overlooked by the window-based method (left) but captured by our method (right). For instance, the syntax-aware attention allows the tokens ``\textit{fresh air}" and ``\textit{allow}" to strongly attend to the token ``\textit{film}", but these tokens are paid less attention in the window-based attention.

\begin{table}[t]
    \centering
    \small
    \begin{tabular}{lcccc}
        \toprule
        & \textbf{QNLI} & \textbf{RTE} & \textbf{MRPC} & \textbf{STS} \\
        Models & Acc & Acc & Acc & PCC \\
        \midrule
        BERT & \textbf{91.7} & \textbf{68.6} & 87.3 & 89.5 \\
        \quad +SLA & 91.4 & 67.8 & \textbf{88.5} & \textbf{89.9} \\
        \bottomrule
    \end{tabular}
    \caption{Experimental results on sentence-pair classification datasets. All models are base-size and results are reported on their dev sets. ``PCC" is short for the Pearson correlation coefficient.}
    \label{tab:paircls}
\end{table}

\paragraph{Testing on Sentence-Pair Classification}
We attempt to evaluate our model on sentence-pair classification datasets. Given a single sentence, we can easily apply dependency parsing and restrain the attention scopes inside the sentence. But for pairwise classification, one problem is how to limit the scopes between a pair of sentences. So a naive approach is adopted, that each token in a sentence can attend to all tokens in another sentence. 
We conduct experiments on four pairwise classification datasets from GLUE benchmark~\citep{wang2018glue}, which cover paraphrase, textual entailment and text similarity.

Experimental results are shown in Table~\ref{tab:paircls}. The syntax-aware local attention achieves better performances on MRPC and STS, but doesn't perform well on RTE and QNLI. We suspect that it is because the cross-sentence interactions are more important for textual entailment task.

\section{Conclusion}
This work verifies that BERT can be further promoted by incorporating syntactic knowledge to the local attention mechanism. With more focused attention over the syntactically relevant words, our model achieves better performance on various benchmarks.
Additionally, the extensive experiments demonstrate the universality of our syntax-aware local attention.


\bibliographystyle{acl_natbib}
\bibliography{main}

\clearpage

\appendix

\section{Appendices}

\subsection{Training Procedure}
\label{sec:train}
We extend the Transformer encoder layer and lift weights from BERT to our model. Following~\citet{bert}, we apply the fine-tuning procedure for various NLP tasks. For classification tasks, the final output of the first token \texttt{[CLS]} is taken as the representation of the input. The probability that the input sentence $X$ is labeled as class $c$ is predicted by a linear transformation with softmax:
\begin{equation}
    P(c~|X) = \text{softmax}(\mathbf{W}_c h_{\texttt{[CLS]}} + b_c)
\end{equation}
where $h_{\texttt{[CLS]}}$ is the representation of the token \texttt{[CLS]}, $\mathbf{W}_c$ and $b_c$ are task-specific parameters. For labeling tasks, we apply the BIO annotation~\citep{bio_scheme} to label outputs and compute the probability that token $x_i$ belongs to class $c$ as:
\begin{equation}
    P(c~|x_i) = \text{softmax}(\mathbf{W}_t h_i + b_t)
\end{equation}
where $h_i$ is the representation of the token $x_i$, $\mathbf{W}_t$ and $b_t$ are task-specific parameters. Finally, the training objective for all tasks is to minimize the cross-entropy loss.

\subsection{Implementation Details}
\label{sec:apd_details}

We apply the whitespace tokenization to the input sentence, and obtain the dependency tree using the Spacy parser\footnote{\url{https://spacy.io/}}. However, the BERT inputs are tokenized by WordPiece tokenizer, which means one word may be split into several sub-words. To address this issue, for each word in the dependency tree, the sub-words split by WordPiece tokenizer share the same masking value in the calculation of syntax-aware local attention.

An important detail is that BERT represents the input by adding a \texttt{[CLS]} token at the beginning as the special classification embedding and separating sentences with a \texttt{[SEP]} token. \citet{bert-look} find that these special tokens are attached with a substantial amount of BERT's attention. Thus, the \texttt{[CLS]} and \texttt{[SEP]} tokens are guaranteed to be present and are never masked out in our local attention.

We use the uncased version of BERT for CoLA and SST-2, and the cased version for CoNLL-2003 and FCE.
During the training, we empirically select the threshold $m$ from \{3,4\}. The maximum sequence length is set to 128 for all tasks. We use Adam~\citep{adam} as our optimizer, and perform grid search over the sets of the learning rate as \{2e-5, 3e-5\} and the number of epochs as \{3,5,10\} for most tasks. In particular, we use smaller learning rates \{5e-6, 1e-5, 2e-5\} and train more epochs \{30, 60\} on CoNLL-2003, but the average F1 of the best 5 runs still hasn't reached the results reported by~\citet{bert}. The batch size is fixed to 32 to reduce the search space, and we evaluate models every 500 training steps for all datasets.
Furthermore, we experiment with the window-based attention on BERT, which allows each token to pay more attention to the neighboring tokens within a window size $2k+1$. We vary the $k$ within \{3,4,5\}, and also incorporate the attention scores with global attention scores.

\begin{table*}[!htb]
    \small
    \centering
    \begin{tabular}{lccccccc}
    \toprule
        ~ & \textbf{ChnSentiCorp} & \multicolumn{3}{c}{\textbf{MSRA NER}} & \multicolumn{3}{c}{\textbf{CGED}} \\
        Models & Acc & P & R & F$_{1}$ & P & R & F$_{1}$ \\
        \midrule
        \multicolumn{8}{l}{\emph{State-of-the-art Models}} \\
        \quad ERNIE 2.0~\cite{ERNIE2} & 95.8 & - & - & 95.0 & - & - & - \\
        \quad BERT-MRC~\cite{mrc-ner} & - & 96.2 & 95.1 & 95.7  & - & - & - \\
        \quad ePMI Matcher~\cite{xunfei_cged}  & - & - & - & - & 83.2 & 61.0 & 70.4 \\
        \midrule
        \multicolumn{8}{l}{\emph{Base-size Models}} \\
        \quad BERT (Our reimplementation) & 94.7 & 95.0 & 94.6 & 94.8 & 79.9 & 75.2 & 77.5 \\
        \quad\quad + WLA & 95.1 & \textbf{95.1} & 94.2 & 94.6 & 79.9 & 73.5 & 76.6  \\
        \quad\quad + SLA & \textbf{95.7} & 94.9 & \textbf{95.0} & \textbf{94.9} & \textbf{81.0} & \textbf{76.6} & \textbf{78.7} \\
    \bottomrule
    \end{tabular}
    \caption{Experimental results on Chinese single-sentence benchmarks. We only show the results of base-size models because Google has not released the large-size model. Reported results are averaged over 5 runs. }
    \label{tab:res_zh}
\end{table*}

\subsection{Testing on Chinese Benchmarks}
The ChnSentiCorp dataset is used for sentiment classification task. We treat the ChnSentiCorp as single-sentence datasets although there are some examples including multiple sentences. The MSRA NER and CGED datasets are selected for named entity recognition and grammatical error detection in Chinese. 
The accuracy (Acc) is used as the metric of ChnSentiCorp, the precision, recall and F$_{1}$ are used as metrics of MSRA NER and CGED. In particular, for a fair comparison with the results of iFLYTEK’s single model~\cite{xunfei_cged}, we construct the CGED test set from CGED 2016 and 2017 test sets. Then we report detection-level results computed by the official evaluation tool.

Table~\ref{tab:res_zh} shows the main results on Chinese datasets. All results are reported on their test set. The proposed syntax-aware local attention outperforms the window-based attention and the basic BERT on all evaluated datasets. We attain 95.7 accuracy on ChnSentiCorp and 94.9 F1 on MSRA NER. Besides, BERT+SLA outperforms the state-of-the-art with a large margin on CGED.

\end{document}